\documentclass[conference]{IEEEtran}
\IEEEoverridecommandlockouts

\usepackage{amsmath,amssymb,amsfonts}
\usepackage{algorithmic}
\usepackage{graphicx}
\usepackage{biblatex}
\usepackage{afterpage}
\usepackage{xcolor}

\usepackage{textcomp}
\usepackage{xcolor}
\def\BibTeX{{\rm B\kern-.05em{\sc i\kern-.025em b}\kern-.08em
    T\kern-.1667em\lower.7ex\hbox{E}\kern-.125emX}}

\begin{document}

\title{Ontology-Based Skill Description Learning for Flexible Production Systems}

\author{\IEEEauthorblockN{ Anna Himmelhuber, Stephan Grimm, Thomas Runkler, Sonja Zillner}
\IEEEauthorblockA{\textit{Siemens AG} \\
Munich, Germany \\
\{anna.himmelhuber, stephan.grimm, thomas.runkler, sonja.zillner\}@siemens.com}
}

\maketitle

\begin{abstract}
 
The increasing importance of resource-efficient production entails that manufacturing companies have to create a more dynamic production environment, with flexible manufacturing machines and processes. To fully utilize this potential of dynamic manufacturing through automatic production planning, formal skill descriptions of the machines are essential. However, generating those skill descriptions in a manual fashion is labor-intensive and requires extensive domain-knowledge. In this contribution an ontology-based semi-automatic skill description system that utilizes production logs and industrial ontologies through inductive logic programming is introduced and benefits and drawbacks of the proposed solution are evaluated. 

\end{abstract}

\begin{IEEEkeywords}
skill description learning, ontology-based, flexible manufacturing, semantic web, inductive logic programming, class expression learning
\end{IEEEkeywords}

\section{Introduction}

In many of today's production facilities, manufacturing machines are deterministically programmed allowing to fulfil one or more predefined tasks. This system works for mass production but cannot address requirements related to flexible manufacturing. Within the Industry 4.0 vision of smart factories, cyber-physical systems are promised to bring more flexibility, adaptability and transparency into production, increasing the autonomy of machines \cite{abele2011zukunft}. In this context, manufacturing processes rely on formal skill descriptions in combination with formalized description of actions related to the single product production requirements as seen in \cite{quiros2018automatic} \cite{palm1990skill}. The term skill refers to the functionalities that a production machine provides. These skill descriptions are the basis for the control functionality of the production process and for fully utilizing the potential of dynamic manufacturing systems \cite{loskyll2012context}\cite{gocev2018explanation}.

To realize cyber-physical systems in production, one approach is to equip machines with explicit digitized skill descriptions, detailing their capabilities.  Having these skill description in a digitized format is necessary for further automation steps like skill matching, where explicit descriptions can be compared to production requests for deciding on the producibility of new product orders and assignment of production modules to production tasks. This method can simplify and speed up the production planning and execution. 
However in some cases, these skill descriptions might not be available at all, e.g. in the case of a legacy module. Even for newer production equipment, skill descriptions, which can contain complex logical structures, might not be available in a digitized format. Without a learning system this has to be done by hand for a variety of existing skill-based approaches as in \cite{hoang2018product}. Defining and digitalizing the skill descriptions of a production module are therefore typically done manually by a domain expert. The domain expert analyzes and conceptualizes the structure of the production process with the respective production modules. Each production module has a specific set of skills and constraints, which must be documented. This process is very labor-intensive and requires a high expertise by the domain expert in order to fully understand the capabilities of a production module. Automatic skill description learning would minimize the labor time and domain expertise needed to equip production modules with their description. 

What is available in most flexible production systems are production logs. These production logs together with industrial ontologies are the basis for and make it possible to learn skill descriptions. Using inductive logic programming (ILP), a sub-field of machine learning that uses first-order logic to represent hypotheses \cite{lehmann2011class} with production logs and ontologies as input, is how we propose to overcome the knowledge acquisition bottleneck for skill descriptions. \\
The contribution of this paper comprises:
\begin{itemize}
    \item Application of state-of-the-art class expression learning for industrial skill descriptions.
    \item Presentation of a skill description learning end-to-end workflow.
    \item Identification of the potential and challenges of using ILP in the skill description learning context.
\end{itemize}

The remainder of the paper is structured as follows: in Section 2, we introduce some notions about skill descriptions and the application of class expression learning. Section 3 introduces the concept of ontology-based skill description learning, the problem formulation and the end-to-end workflow and architecture for skill description generation we have developed. In Section 4, we describe and evaluate the results of the experiments we have conducted. Section 5 presents the conclusions of our contribution. 

\begin{figure*}[ht]
\includegraphics{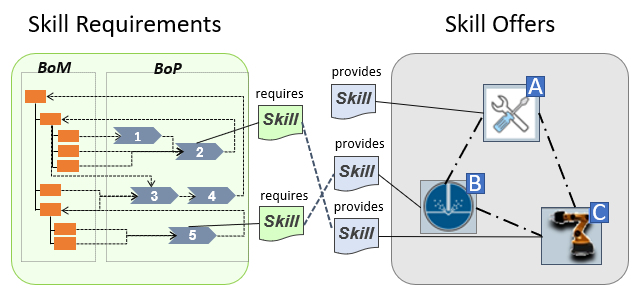}
 \centering
\caption{Skill Matching}
\label{skillmatching}
\label{fig}
\end{figure*}

\section{State of the Art}

\subsection{Skill Descriptions}
Since skill descriptions are crucial to make a dynamic production environment possible,  a number of approaches have been proposed for the modelling of functionalities in a manufacturing environment. In literature, concepts for skill descriptions, tailored to specific process-oriented needs, are introduced. 
In \cite{jarvenpaa2016formal} skill functionalities are comprised by its name and parameters, which provides the possibility of combining several simpler functionalities to more complex ones. The objective of the skill descriptions is matching verification, in which certain product requirements should be met by the production module.
In  \cite{pfrommer2015plug}, the approach is based on the notion that a bill of processes (BoP) for a product is available. This BoP is then matched against the functionality description of the production module. The functionality, e.g. "drilling" is described as the ability of the production module to execute a specific production process and is constrained by attributes like "depth"  with their values. 
Another product-oriented concept \cite{hoang2018product} concentrates on facilitating the feasibility check of product requirements, i.e. checking if a request can be fulfilled by a resource and if the production planning time could be reduced.
Approaches like DEVEKOS  use skill descriptions not only in the phase of engineering, but also consequently for direct and generic control of field-devices. An executable skill-metamodel therefore describes the methodological functionality as well as built-in services and the information model. \cite{8869473}

\subsection{Class Expression Learning}

Semantic technologies can provide formal description and semantic processing of data, therefore making the data interpretable with regard to its content and meaning.
This explicit knowledge representation of the Semantic Web includes modeling of knowledge and application of formal logics over the knowledge base. One approach are ontologies, which enable the modeling of information and consist of classes, relations and instances \cite{loskyll2012context}. 
 Class expression learning (CEL) is a sub-field of inductive logic programming, where a set of positive and negative examples of individuals are given in an ontology. The supervised learning problem consists of finding a new class expression, such that most of the positive examples are instances of that concept, while the negatives examples are not \cite{lehmann2011class}.  CEL and ILP will be used interchangeably in this paper.  

In literature, a few application of CEL for solving different problems have been proposed. In Sentiment Analysis \cite{salguero2016description}, CEL is introduced as a tool to find the class expression that describes as much of the instances of positive documents as possible. Here, the ontology is focused on inferring knowledge at a syntactic level to determine the orientation of opinion.
Another application example is the use of class expression learning for the recognition of activities of daily living in a Smart Environments setting \cite{salguero2019ontology}.

To the best of our knowledge, class expression learning has not been used for learning skill descriptions in a manufacturing setting.



\section{Learning Ontology-Based Skill Descriptions}

\subsection{Production Environment}
\label{prod}

The type of flexible production system we are looking at 
consists of one or more production lines with a number of production modules. These production modules or machines, have a set of skills, for which we want to learn descriptions. 
The production orders consist of a bill of materials (BoM) and a bill of processes which are used for the automatic production planning - production steps are assigned to specific production modules and scheduled for a certain time within a specific production plan. This enables an efficient production process and allocation of resources. Part of this process is skill matching (see Figure \ref{skillmatching}), where the skill requirements of a certain operation are matched to the skill offers of a production module. For example, manufacturing process step two, requires an intermediate product, created in step one and one more material as seen in the BoM and BoP. These two parts have to be joined which requires a joining skill of a production module. The production module $C$ offers this skill, and the skill requirement and skill offer can be matched. However, this requires the skill offer of module $C$ to be available in a digital format, to make a successful skill matching possible. \\

Figure \ref{ontology} shows a screenshot of an ontology detailing an assembly scenario in the Protégé application \cite{DBLP:journals/aimatters/Musen15}. In order to exemplify the skill description learning, we chose one example skill carried out by a specific module: \emph{Assembling of an Item by Module 1}. 
 Here, one item or material is assembled onto another item by the production module \emph{Module1}. Other skills that could be looked at  include joining, charging dismantling, recycling, etc.  In the ontology we can see: 

\begin{figure*}[ht]
\includegraphics[width=19cm]{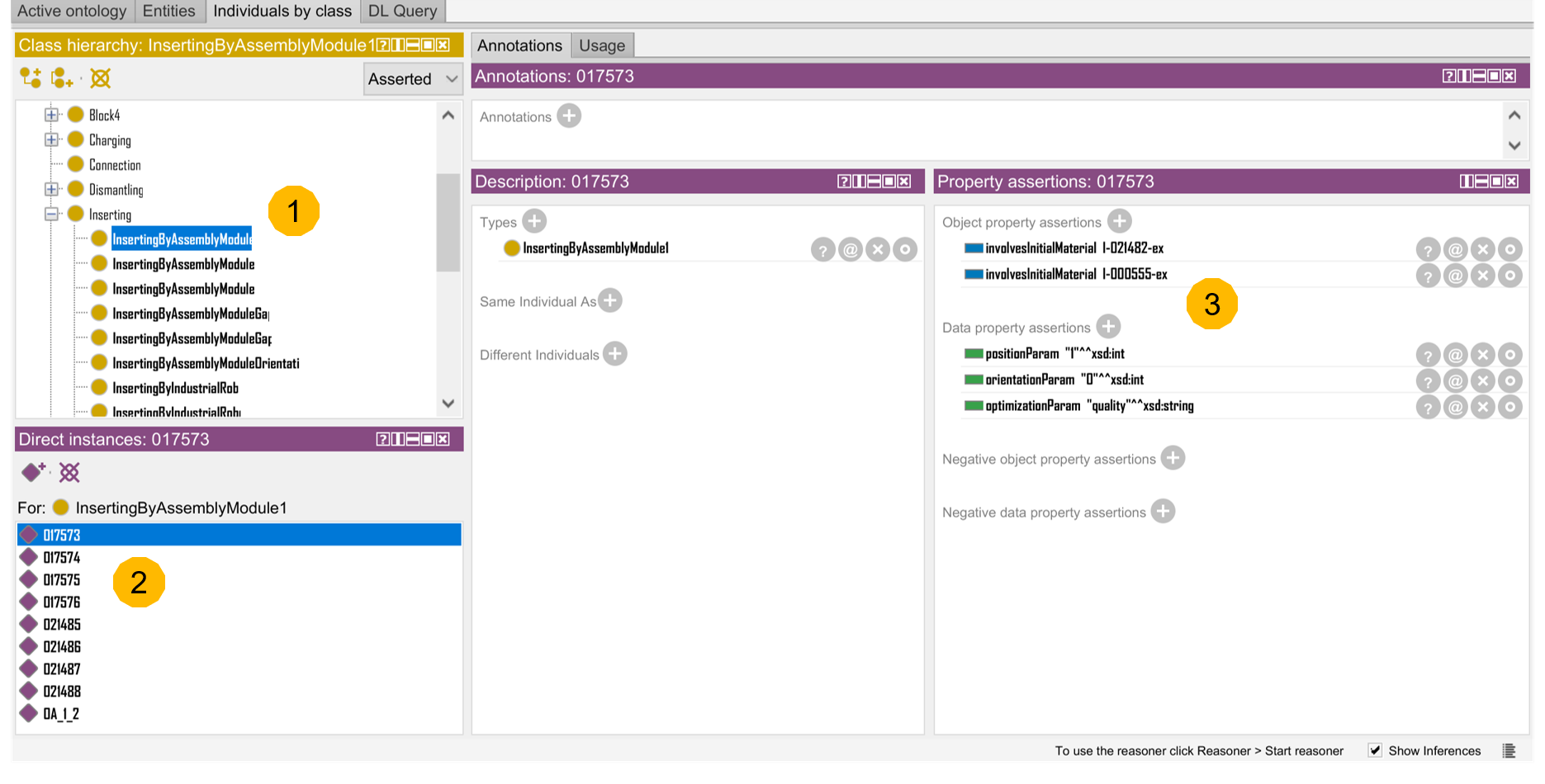}
\caption{Ontology about an Assembly Scenario}
\label{ontology}
\end{figure*}

\begin{enumerate}
    \item The class hierarchy of all the production modules, materials, etc. All classes that are relevant to the production process are modelled here. 
    \item The example data used by  ILP are the instances of the operation carried out by the module. These are the production logs, modelled in the ontology. 

    \item The object properties are the background knowledge of each single operation instance. These are properties or constraints of the skill descriptions we want to learn. The properties are used to assert relationships between individuals, in this case, the operation instance \emph{I-017573-ex} has the object property Position Parameter \emph{PositionParam} of "Position 1".  Another instance of this operation has the Position Parameter of "Position 2". Therefore, our algorithm should find a class expressions, that expresses that the position parameter of our operation has Position Parameter "Position 1" or "Position 2".  

\end{enumerate}

The ground truth for this skill description for skill \emph{AssembleItemByModule1} example is comprised of three "class expression"-like statements or constraints and is generated manually by a domain expert as OWL constraints:
\begin{itemize}
    \item Material involved has to be \emph{MaterialProductBase} or \emph{BottomPart}
    \item Object has Position Parameter \emph{Position 1} or \emph{Position 2}
    \item Object has Orientation Parameter \emph{hundredeighty} or \emph{zero} 
\end{itemize}

\subsection{Description Learning Problem Formulation}
We represent log data of the production processes as instance data $I$ and ontologies as modelled background data $B$ containing machine and product parameters, with instance data $I$ and background data $B$ constituting knowledge base $K$. $A_{total}$ represents the ground truth skill description and is a set of class expression interpreted as a conjunction of the respective constraints and properties of the skill. These are represented in the form of OWL restrictions on properties.
Similarly, the skill description $A_{learned}$ is a set of learned class expressions $A_i$, with
\[ A_{learned} = \{A_1, ...,A_n\},\]where each class expression $A_i$ represents a constraint or a property of the skill. $A_{learned}$ is a subset of $C$, with $C$ being a list of all possible class expressions $C_i$ for a production module created by inductive logic programming. In the next step a domain expert can decide which class expressions $C_i$ are most appropriate, based on key indicators. The set of selected class expressions $A_{learned}$ constitutes a skill description. For a complete and concise learned skill description
\[ A_{learned} = A_{total}\]
should apply. 
The data used for learning the class expressions $C_i$ is captured by semantic web technology, more specifically by ontologies describing cyber-physical systems.  This background knowledge represent domain knowledge about the equipment of a production plant, products and their production requirements, materials and production processes. 

\subsection{Workflow and Architecture of end-to-end Skill Description Generation}\label{CC}

\begin{figure*}[ht]
\centering
\includegraphics[width=18cm,height=10cm]{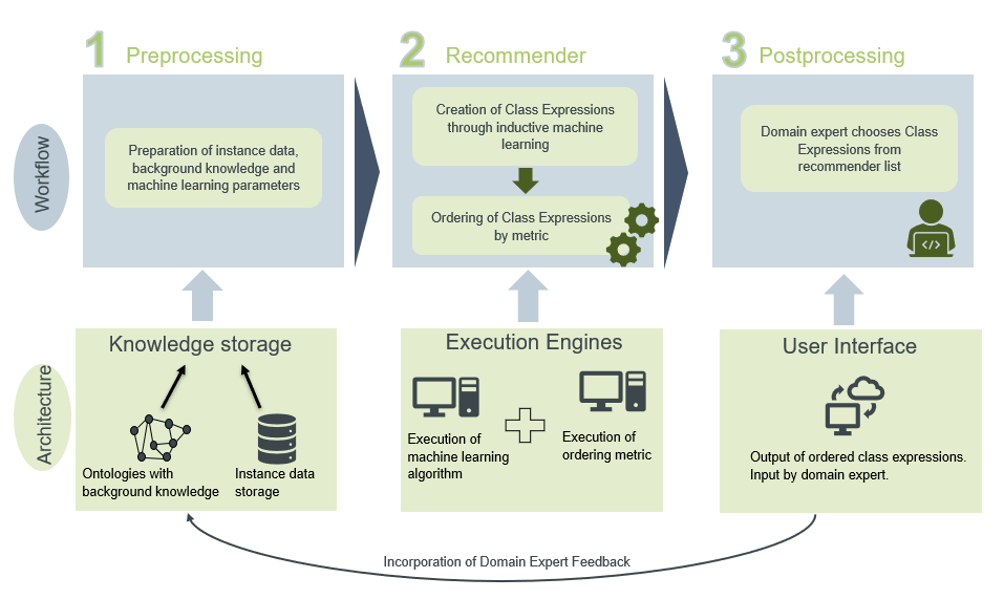}
\caption{Process Workflow and Architecture}
\label{workflow}
\end{figure*}

The workflow of our skill description learning system can be subdivided into three building blocks as seen in Figure \ref{workflow}. It includes the preprocessing, recommender and postprocessing building blocks:

\begin{enumerate}
\item The preprocessing building block contains the preparation of the example data $I$, which is resulting from the log data. Each example $I_i$ is an individual in our knowledge base $K$, i.e. an operation carried out by the specific production module as can be seen in Figure \ref{log}. Information captured by the log data include the operation ID, the machine carrying out the operation, the skill name and the operation duration. 

In order to achieve meaningful class expressions, the individuals in the ontology need to be equipped with background knowledge. An example for background knowledge would be information detailing the operation in the production process, such as the material involved as seen in Figure \ref{rdf}. The learned class expression given by the class expression learner, has OWL Manchester Syntax:\\

\emph{involvesMaterial only (MaterialProductBase or BottomPart1)}\\

The Manchester OWL syntax is a user-friendly syntax for OWL Description Logics, fundamentally based on collecting all information about a  particular class, property, or individual into a single construct \cite{horridge2008manchester}.
This background knowledge is modelled in an ontology as seen in Figure \ref{ontology}. 
For a successful class expression learning, a high quality of the ontology is needed. Modelling errors, i.e. missing or wrongly assigned background knowledge, can lead to a reduced quality of the final skill descriptions. For example an operation instance assigned to the wrong skill name could lead to erroneous class expressions.

\begin{figure}[ht]
\centering
\includegraphics{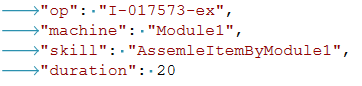}
\caption{Instance Data: Example Operation}
\label{log}
\end{figure}

\begin{figure}[ht]
\includegraphics{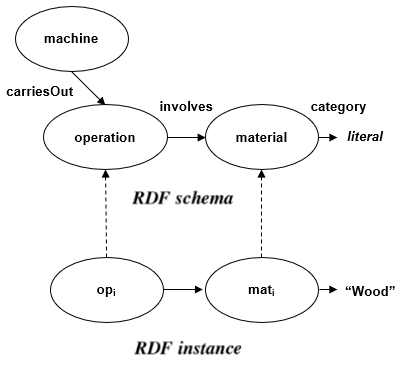}
\caption{RDF Schema and Instance Example}
\label{rdf}
\end{figure}

\item The recommender building block contains two steps. Firstly, the machine learning part of the system. Inductive logic programming is a search process, which takes operations carried out by the production module we want to describe as positive examples $I$ and creates and tests class expressions $C$ against a background knowledge base $B$. 
Secondly, in order to keep our approach efficient, the collection of most fitting class expressions should be found high up on the recommender list, as the domain expert can only go through and evaluate a limited number of suggestions. The ordering of the class expressions is done by predictive accuracy.
We have implemented the algorithm within the open-source framework DL-Learner, which can be used to learn classes in OWL ontologies from selected objects. It extends ILP to Descriptions Logics and the Semantic Web \cite{buhmann2016dl}.
Based on existing instances of an OWL class, the DL-Learner can make suggestions i.e. generate class expressions for class descriptions. In our example in Figure \ref{ontology}, the instances (2) from \emph{Operation1}, subclass to \emph{AssembleItemByModule1}, are the basis for its class description. The standard algorithm for class expression learning, namely \emph{CELOE}, was used.

\item The postprocessing building block involves a domain expert, who selects the class expressions given by the recommender according to a set of predefined key indicators including completeness, accuracy and human-understandability.  The final skill description $A_{learned}$ is saved to the knowledge storage and can then be used in further flexible manufacturing processes like skill matching. 


\end{enumerate}

The architecture of our approach includes a knowledge storage for ontologies holding the background knowledge and instance data, an execution engine for executing ILP and a user interface for the domain expert to interact with, as can be seen in Figure \ref{workflow}.

\section{Evaluation}

In this section we evaluate the results of the recommender building block outlined above, so the quality of the class expressions generated from the DL-Learner with production ontologies as background data and production logs as positive examples for a skill. We limit the DL-Learner results to a list of the top 20 results to uphold the efficiency of the approach. Since the post-processing step includes a domain expert selecting the wanted class expressions, we need to limit the potential choices to a reasonable amount. This is a efficiency versus completeness trade-off, since ground truth class expressions could fall outside of the top 20 results. These are ordered by predictive accuracy since in standard applications of the DL-Learner, the primary purpose is to find a class expression, which can classify unseen individuals, i.e. not belonging to the examples, correctly.
Predictive accuracy is measured as the number of correctly classified examples divided by the number of all examples \cite{lehmann2011class}.
However, it is not the best choice for this use case since we don't use the skill descriptions for classification but production planning.
The ideal output would be ordered according to a completeness aspect: We want a combination of class expressions that gives us a complete and precise description of a certain production module skill. This means that all constraints and properties of a skill should be described in a concise manner. Therefore, the metrics recall and precision are used for the evaluation. 
\subsection{Qualitative Evaluation}\label{AA}
In Table \ref{rec} you can see an example of the recommender list result for the \emph{AssembleItemByModule1} skill. 
The class expressions number 1, 2, and 18 are the ground truth (as stated in section \ref{prod}) and can all be found in the top 20 results.
However, some of the other class expressions have very little or no useful information.  
For example class expression number 5 \emph{involvesMaterial max 1 PartType3} isn't wrong, in that no material of type \emph{PartType3} is used in this skill. But including this class expression in the skill descriptions  wouldn't add any value to a concise and complete description and could diminish skill description understandability. That is why a domain expert is still needed, to discern between the useful and useless class expressions to generate a complete skill description. To do so, the domain expert has to evaluate all 20 class expressions and choose a subset based on their content and style for the final skill description.

\begin{table}[ht]
\centering
\begin{tabular}{ |p{0.3cm}||p{6cm}|p{1.3cm}|  }
 \hline
 \multicolumn{3}{|c|}{
 Class Expressions
 } \\
 \hline
 \#     & Class Expression & Pred. Acc.\\
 \hline
1. & \emph{involvesMaterial only (MaterialProductBase or BottomPart)} & 100.00\%\\
2. & \emph{hasPositionParam only (pos1 or pos2)} &100.00\%\\
3. & \emph{Thing} & 40.82\%\\
4. & \emph{involvesMaterial max 1 Thing} &  40.82\%\\
5. & \emph{involvesMaterial max 1 PartType3} & 40.82\%\\
6. & \emph{involvesMaterial max 1 PartType2} & 40.82\%\\
7. & \emph{involvesMaterial max 1 PartType1} & 40.82\%\\
8. & \emph{involvesMaterial max 1 HeadPart} & 40.82\%\\
9. & \emph{involvesMaterial max 1 BottomPart} & 40.82\%\\
10. & \emph{involvesMaterial max 1 MaterialProductBase} & 40.82\%\\
11. & \emph{involvesMaterial max 1 zero} & 40.82\%\\
12. & \emph{involvesMaterial max 1 pos5} & 40.82\%\\
13. & \emph{involvesMaterial max 1 pos4} & 40.82\%\\
14. & \emph{involvesMaterial max 1 pos3} & 40.82\%\\
15. & \emph{involvesMaterial max 1 pos2} & 40.82\%\\
16. & \emph{involvesMaterial max 1 pos1} & 40.82\%\\
17. & \emph{involvesMaterial max 1 hundredeighty} & 40.82\%\\
18. & \emph{hasOrientationParam only (hundredeighty or zero)} & 40.82\%\\
19. & \emph{pos1 or (involvesMaterial max 1 Thing)} & 40.82\%\\
20. & \emph{hundredeighty or (involvesMaterial max 1 Thing)} & 40.82\%\\

 \hline

\end{tabular}
\newline
\caption{Class Expression Recommender List}
\label{rec}
\end{table}

\subsection{Quantitative Evaluation}
Experiments were carried out for four different skills, which show some variability in terms of constraints and properties: \emph{ AssembleItemByModule1,  AssembleItemByModule2,  DismantleProductByModule3} and \emph{ChargeProductBaseByModule4}.

In order to evaluate the class expressions results, we define the calculations for the True Positives, False Negatives and False Positives. True Negatives don't play a role and cannot be calculated, since they are the class expressions that aren't in the ground truth and haven't been generated. 
\begin{itemize}
    \item  \emph{True Positives = $|A_{learned}|$ = number class expressions found in top 20 results and ground truth},
    \item \emph{False Negatives = $|A_{total}|-|A_{learned}|$ = number of class expressions not found in top 20 results, but found in ground truth} and 
    \item \emph{False Positives = $|C|-|A_{learned}|$ = number of class expressions found in top 20 results, but not found in ground truth}

\end{itemize}

We used following metrics to evaluate the quality of the recommender list results:
\begin{enumerate}
\item Recall : How many of the ground truth class expression are found in the top 20 class expressions?
\begin{multline}
Recall = \frac{|A_{learned}|}{|A_{learned}| + (|A_{total}|-|A_{learned}|) } \\
= \frac{|A_{learned}|}{|A_{total}| }
\end{multline}
\item Precision: How many of the top 20 class expressions are in the ground truth class expressions?
\begin{multline}
Precision = \frac{|A_{learned}|}{|C|-|A_{learned}|+|A_{learned}|} = \\ \frac{|A_{learned}|}{|C|} = \frac{|A_{learned}|}{20}
\end{multline}
\end{enumerate}

The recall gives us is the fraction of the total amount of relevant class expressions that were actually retrieved. Can all the wanted class expressions be found in a reasonable long list of class expressions?
The precision is the  fraction of relevant class expressions among the retrieved class expressions. Does the domain expert have to deal with lot of false positives? \\

The evaluation results are shown in Table \ref{table}. As can be seen, all relevant class expressions are found for three out of four skills, with Recall = 1. This means, that in three out four cases,  no additional manual class expression generation is needed to arrive at a complete skill description. 
The precision is relatively low with 10\%-15\%. This stems from the skills having only a total of 2-3 ground truth statements that could be found in the top 20 class expressions. Would all generated class expression be included in the skill description, we would have a convoluted and imprecise result. Therefore the additional postprocessing selection step is warranted. Notice, however, that even a precision as low as 0.1 means that on average two out of twenty found classes are correct, so this approach is considered very useful in industrial applications, and selecting the correct classes from the automatically found suggestions is a much lower effort than manual labeling from scratch. The results show, that our approach is a valid alternative to crafting skill descriptions manually, with significantly less labor time and domain-expertise needs. 
\\

\begin{table}[ht]
\centering
\begin{tabular}{ |p{4cm}||p{1cm}|p{1cm}|  }
 \hline
 \multicolumn{3}{|c|}{Skill Descriptions} \\
 \hline
 skill     & Recall & Precision\\
 \hline
 AssembleItemByModule1   & 1    &0.15\\
 AssembleItemByModule2   & 0.67    &0.10\\
 DismantleProductByModule3   & 1    &0.10\\
 ChargeProductBaseByModule4   & 1    &0.15\\

 \hline

\end{tabular}
\newline
\caption{Skill Description Recall and Precision}
\label{table}
\end{table}

\section{Conclusion}

This contribution describes how class expression learning can be applied to the field of production module skill descriptions. 
It demonstrates that learning skill descriptions with ILP decreases labor and domain expertise needs, even with low precision scores. However, ILP-based learning should not be seen as an stand-alone approach, but a building block in a workflow which also includes preprocessing and postprocessing building blocks. 
Disadvantages of the approach include the ontology quality requirements. Errors in the ontology modelings might lead to a reduced quality of class expressions results. However, in setups with fully-automated skill matching, it can be assumed that ontologies have to have a certain level of quality as otherwise the skill matching wouldn't work in the first place. Also, the typically available production logs can be exploited, which helps the preprocessing building block.  Since the skill descriptions are generated from the log data,  skill descriptions or offers and skill requirements utilize the same semantics, which can facilitate feasibility checks. \\

As possible future work one can mention an implementation of a more to skill description learning adapted algorithm and ordering, where recall and precision are maximised and therefore the domain expert effort is reduced.

\newpage

\section*{Acknowledgements}
This work was supported by the German Federal Ministry
of Economics and Technology (BMWI) in the
project RAKI (no. 01MD19012D).

\printbibliography

\vspace{12pt}

\end{document}